\documentclass{./styles/svproc}
\usepackage{url}
\usepackage{graphicx}
\usepackage{xcolor}

\usepackage{amsmath} 
\usepackage{amssymb}
\usepackage{url}
\usepackage{hyperref}

\begin{document}
\mainmatter              
\title{CBF-Based STL Motion Planning for Social Navigation in Crowded Environment\thanks{This work was supported by Horizon Europe program under the Grant Agreement 101070351 (SERMAS).
}}
\titlerunning{CBF-Based STL Motion Planning}  
%
\author{Andrea Ruo \and Lorenzo Sabattini \and Valeria Villani}

\authorrunning{Andrea Ruo et al.} 
%
\tocauthor{Andrea Ruo, Lorenzo Sabattini, and Valeria Villani}

\institute{{University of Modena and Reggio Emilia, Reggio Emilia RE 42122, ITALY}\\
\email{\{name.surname\}@unimore.it}}

\maketitle              
\vspace{-.5cm}
\begin{abstract}
A motion planning methodology based on the combination of Control Barrier Functions (CBF) and Signal Temporal Logic (STL) is employed in this paper. 
This methodology allows task completion at any point within a specified time interval, considering a dynamic system subject to velocity constraints. In this work, we apply this approach into the context of Socially Responsible Navigation (SRN), introducing a rotation constraint. 
This constraint is designed to maintain the user within the robot's field of view (FOV), enhancing human-robot interaction with the concept of side-by-side human-robot companion. This angular constraint offers the possibility to customize social navigation to specific needs, thereby enabling safe SRN. Its validation is carried out through simulations demonstrating the system's effectiveness in adhering to spatio-temporal constraints, including those related to robot velocity, rotation, and the presence of static and dynamic obstacles.

\end{abstract}
\section{Introduction}
\label{sec:introduction}
In the recent years, an increasing number of robots have entered human environments. To navigate in these places, the robot needs to be aware of the humans in the environment, and treating humans simply as obstacles may not be enough. Furthermore, the robot's motion should be safe, legible and acceptable to humans rather than being optimal from the sole point of view of the robot \cite{HumanAwareNavigationPlanner}. In SRN context, a mobile robot operates autonomously within an environment, facing the challenge of navigating a path successfully while avoiding collisions with obstacles in the environment \cite{LidarSlam}. Certain social spaces can be wide and crowded. In these scenarios, it becomes crucial for robots to move exhibiting appropriate social behaviors \cite{onlineRobotNavigation}, such as the side-by-side human-robot companion. Specifically, when guiding the user to their destination, the robot can stand next to or in front of the human, while moving forward. This factor can significantly impact the acceptance of human-robot interaction.

Recent literature has explored similar concepts in the context of SRN, investigating various approaches and examples. The SPENCER project, funded by the European Union \cite{Spencer}, has developed a reception robot tailored to assist, inform, and guide passengers in large and crowded airports. This robot combines map representation, laser-based people and group tracking, and activity and motion planning. Stricker et al. \cite{university} proposed a robot-based information system for a university building. The reception robot provides information about offices, employees, and laboratories in the building and can guide visitors to their desired locations. In the future, we expect to see social robots sharing urban areas with people. To achieve this integration, robots have to develop several skills, including the ability to navigate alongside with humans \cite{SidebySideCompanion}. Research on human-robot side-by-side navigation is relatively new compared to traditional robot navigation, in which robots navigate in a safe and human-like manner.

In the given application context, it is essential to include safety-related constraints, including obstacle avoidance, velocity limits, and speed reduction when the robot is in close proximity to people. Furthermore, space-time constraints might be relevant to guarantee the efficient and safe execution of activities, particularly social environments. Temporal constraints can appear in various manifestations, such as time limits to complete a specific task, time intervals to complete a sequence of tasks, or priorities assigned to different activities based on their importance. These constraints might be stipulated by environmental necessities or user preferences.
    
To address and implement the discussed constraints, tools such as STL and CBF can be valuable. Temporal logics, such as STL \cite{maler2004monitoring}, enable the specification of spatio-temporal constraints, enhancing the expressiveness of Boolean logic by incorporating the temporal dimension. This allows the use of specific expressions as constraints, such as ``\textit{the robot must reach the goal pose within 10 seconds}''. The relationship between the semantics of an STL task and time-varying CBFs enables the formal control of systems, ensuring compliance with spatio-temporal constraints and maintaining safety.


This paper focuses on advancing the CBF-based STL motion planning methodology, first presented in~\cite{ruo2024cbf}. 
The methodology, originally designed for general motion planning with temporal constraints, is now applied in the domain of SRN. In doing so, we introduce and validate a angular constraint. This addition is aimed at enhancing human-robot interaction by ensuring the user remains within the robot's FOV. This contribution enables personalized social navigation, especially when the robot serves as the user's companion, ensuring visual contact to enhance engagement and safety in SRN.

\section{Preliminaries and Problem Statement}
\label{sec:preliminaries}
Let $\boldsymbol{x} \in \mathbb{R}^n$ and $\boldsymbol{u} \in \mathcal{U} \subseteq \mathbb{R}^m$ be the state and input of a nonlinear input-affine control system:
\begin{equation} 
    \dot {\boldsymbol {x}}=f(\boldsymbol {x})+g(\boldsymbol {x})\boldsymbol {u}
    \label{eq:modelSystem}
\end{equation}

Referring to the work in \cite{ruo2024cbf}, the problem under consideration in this paper can be stated as follows.
\begin{problem}
\label{prob:prob1}
   \emph{Given the dynamical system in~\eqref{eq:modelSystem} and an STL fragment $\phi$ \cite{lindemann2020barrier}, derive a control law $\boldsymbol{u}(t)$ so that the solution $\boldsymbol{x}:\mathbb{R}_{\geq 0} \rightarrow \mathbb{R}^n$ to (\ref{eq:modelSystem}) is such that $(\boldsymbol{x},0)$ satisfies $\phi$ providing safety-critical guarantees regarding non-linear velocity constraints, angular constraint and obstacle avoidance.}
\end{problem}


\section{Approach}
\label{sec:Implementation}
\begin{figure}
      \centering
      \includegraphics[width=0.55\textwidth]{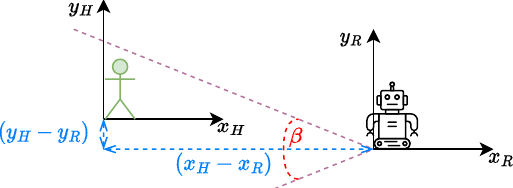}
      \caption{Angular constraint.}
      \label{img:angular}
\end{figure}

In this paper we focus on the need of keeping the user in
a limited angular sector, with amplitude $\beta$, in order to be detected by the robot, as shown in Fig.~\ref{img:angular}.

A CBF is defined starting from the relative position of the detected user $H$ respect to the robot $R$, expressed as ${^R\boldsymbol{p}_H}=[x_H-x_R,y_H-y_R]^T$, with two separates components, one from the left boundary of the FOV (i.e., the lower dashed line in Fig.~\ref{img:angular}) and one from the right boundary (i.e., the upper dashed line in Fig.~\ref{img:angular})\cite{NeighbourVisionDetection}. The two components are treated as different constraints inserted into an optimization solver. To this end, we introduce the following $h(\cdot)$ for the robot:
\begin{equation}
    h({^R\boldsymbol{p}_H})= \begin{bmatrix}   h_1({^R\boldsymbol{p}_H})   \\[0.3em] h_2({^R\boldsymbol{p}_H}) \end{bmatrix} = - \begin{bmatrix}  tan(\frac{\beta}{2}) & 1  \\[0.3em] tan(\frac{\beta}{2}) & -1 \end{bmatrix} {{^R\boldsymbol{p}_H}}.
    \label{eq:barrier_angularConstraint}
\end{equation}
Imposing the condition $h(\cdot)\geq \boldsymbol{0}$ is equivalent to constraining the robot's orientation such that the user is in the angular sector $\beta$, as shown in Fig.~\ref{img:angular}. The time derivative of this formulation is then expressed as
\begin{equation}
    \dot h({^R\boldsymbol{p}_H}, \boldsymbol{\hat{u}})= \frac{\partial h({^R\boldsymbol{p}_H})}{\partial {^R\boldsymbol{p}_H}}{^R\boldsymbol{\dot p}_H}= -\begin{bmatrix}  tan(\frac{\beta}{2}) & 1  \\[0.3em] tan(\frac{\beta}{2}) & -1 \end{bmatrix} {^R\boldsymbol{\dot p}_H}.
    \label{eq:derivative_barrier_angularConstraint}
\end{equation}
The velocity of $H$ with respect to $R$ is then expressed through kinematic computations: 
\begin{equation}
    {^R\boldsymbol{\dot p}_H}= {^RR_W}{^W\boldsymbol{\dot p}_H}-\begin{bmatrix}  {^Rv_x}  \\[0.3em] {^Rv_y} \end{bmatrix} + \omega \begin{bmatrix}  y_H-y_R  \\[0.3em] -(x_H-x_R)
    \end{bmatrix}.
\end{equation}
%
To solve Problem~\ref{prob:prob1}, it is possible to integrate a dedicated CBF that enables the execution of a side-by-side human-robot companion in the quadratic problem described in \cite{ruo2024cbf}, obtaining the formulation expressed in \eqref{eq:qp_2}.
In conclusion, we can obtain $\boldsymbol{u}(\boldsymbol{x},t)$ as 
\begin{equation}
\label{eq:qp_2}
\resizebox{\hsize}{!}{
    $
    \begin{aligned}
        &\underset{\hat{\boldsymbol{u}}\in\mathcal{U}}{\operatorname{min}}\; \hat{\boldsymbol{u}}^T Q\hat{\boldsymbol{u}},\\
        \text{s.t.}\; &\frac{\partial {\mathfrak{b}}({\boldsymbol{x}},t)}{\partial \boldsymbol{x}} f(\boldsymbol{x})+g(\boldsymbol{x})\hat{\boldsymbol{u}}+\frac{\partial {\mathfrak{b}}({\boldsymbol{x}},t)}{\partial t}\geq-\alpha(\mathfrak{b}({\boldsymbol{x}},t)), \\
        &\left \|v_x + v_y\right \| \leq v_{max}, \\
        &\frac{\partial h({^R\boldsymbol{p}_H})}{\partial {^R\boldsymbol{p}_H}} {^RR_W}{^W\boldsymbol{\dot p}_H} - \frac{\partial h({^R\boldsymbol{p}_H})}{\partial {^R\boldsymbol{p}_H}} \begin{bmatrix}  {^Rv_x}  \\[0.3em] {^Rv_y} \end{bmatrix} + \frac{\partial h({^R\boldsymbol{p}_H})}{\partial {^R\boldsymbol{p}_H}} \omega \begin{bmatrix}  y_H-y_R  \\[0.3em] -x_H+x_R \end{bmatrix} \geq - \alpha (h({^R\boldsymbol{p}_H})), 
    \end{aligned}
    $
}
\end{equation}
where the first constraint formulates the CBF-STL constraint to ensure the completion of a series of tasks within specified time intervals while ensuring obstacle avoidance. The second constraint imposes that the squared norm of the non-linear velocity is less than or equal to $v_{max}$.
The third constraint guarantees that the user is always in the robot's FOV.

\section{Simulation Results}\label{sec:simulation}
\begin{figure}[thpb]
      \centering
      \includegraphics[width=.95\textwidth]{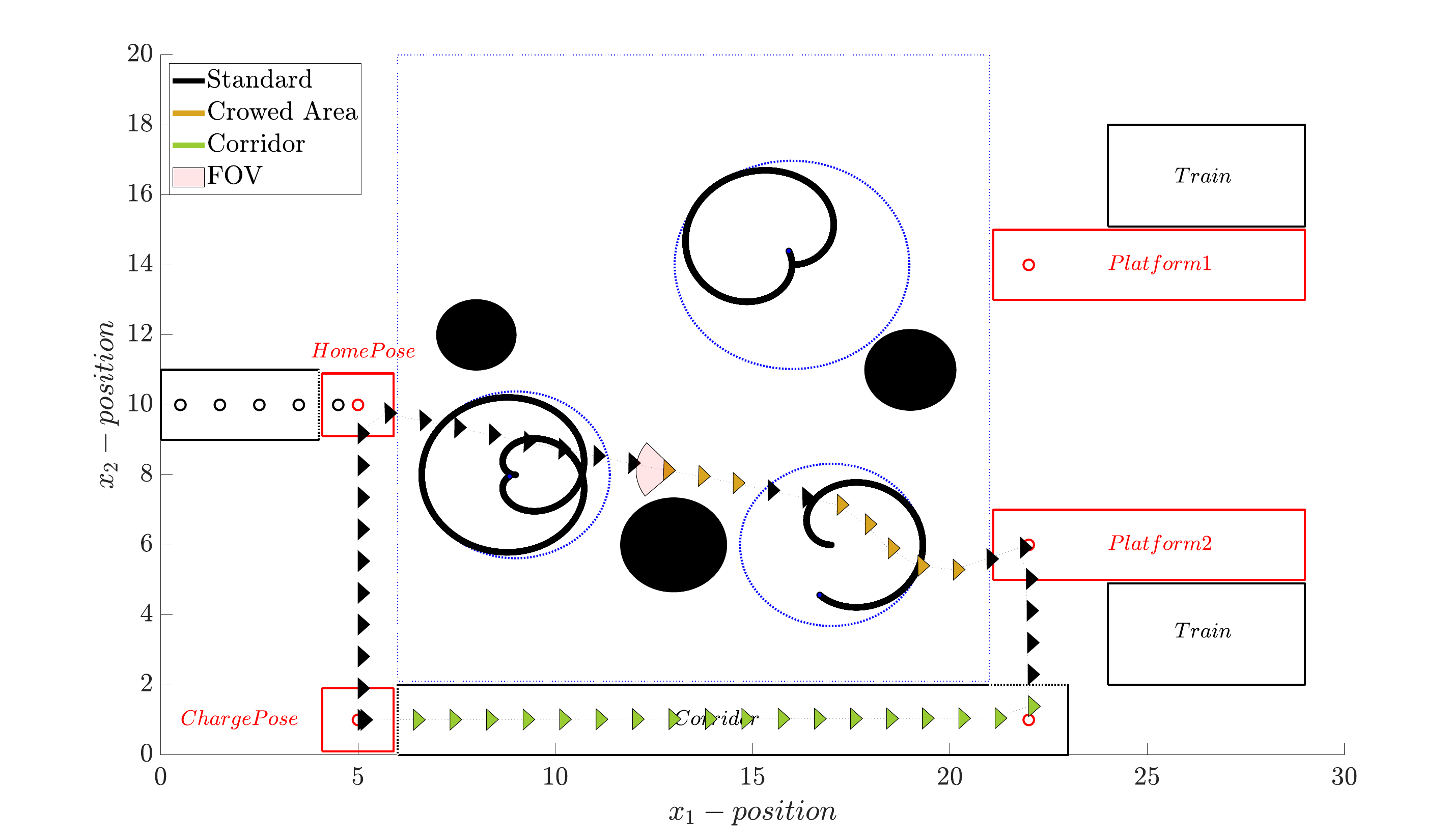}
      \caption{The robot starts from \emph{ChargePose} and, upon detecting the user, moves to \emph{HomePose}. The robot guides the user to the \emph{Platform2} while ensuring velocity constraints, angular constraints, and collision avoidance. Subsequently, the robot returns to the home pose through an obstacle-free corridor.}
      \label{img:simulation}
\end{figure}

\vspace{-2mm}
We simulated an SRN scenario in Matlab, shown in Fig.~\ref{img:simulation} and in the accompanying video\footnote{\href{https://doi.org/10.5281/zenodo.10255241}{DOI: https://doi.org/10.5281/zenodo.10255241}}, using a similar simulation proposed in \cite{ruo2024cbf}. The input control $\boldsymbol{u}$ are the angular velocities of the robot, considering the three-wheeled omnidirectional robot model described in \cite{liu2008omni}. We assumed maintaining the user at an arbitrary distance with a certain level of noise from the robot.

Using the proposed approach, it is possible to observe the results of motion planning, which allows for the identification of a valid path for the robot and the satisfaction of STL specifications subject to non-linear velocity constraints, angular constraint, ensuring the compliance with safety guarantees.

\section{Conclusion}\label{sec:conclusion}
\vspace{-2mm}
In this work, we introduced an extension to our prior research, incorporating a new constraint to facilitate side-by-side interaction between humans and robots. We presented a simulation to validate the proposed approach. As a further development, we are going to integrate prediction methods to enable the robot to adjust its motion based on the behavior of the person being accompanied.

%
%
\bibliographystyle{unsrt}
\bibliography{main}
\end{document}